\documentclass{article}

\usepackage[preprint]{corl_2024} 

\usepackage{stackengine}
\usepackage{appendix}
\usepackage{graphicx}
\usepackage{array}
\usepackage{caption}
\usepackage{subcaption}
\usepackage{amsmath}
\usepackage{algorithm}
\usepackage{algpseudocode}

\usepackage[many]{tcolorbox}
\usepackage{xcolor}
\usepackage{varwidth}
\usepackage{environ}
\usepackage{xparse}

\usepackage[T1]{fontenc}

\newlength{\bubblewidth}
\AtBeginDocument{\setlength{\bubblewidth}{.75\textwidth}}
\definecolor{bubblegreen}{RGB}{103,184,104}
\definecolor{bubblegray}{RGB}{241,240,240}

\newcommand{\bubble}[4]{%
  \scriptsize\tcbox[
    colback=#1,
    colframe=#1,
    #2,
  ]{\color{#3}\begin{varwidth}{\bubblewidth}#4\end{varwidth}}%
}

\ExplSyntaxOn
\seq_new:N \l__ooker_bubbles_seq
\tl_new:N \l__ooker_bubbles_first_tl
\tl_new:N \l__ooker_bubbles_last_tl

\NewEnviron{rightbubbles}
 {
  \raggedleft\sffamily
  \seq_set_split:NnV \l__ooker_bubbles_seq { \par } \BODY
  \int_compare:nTF { \seq_count:N \l__ooker_bubbles_seq < 2 }
   {
    \bubble{bubblegreen}{rounded~corners}{white}{\BODY}
   }
   {
    \seq_pop_left:NN \l__ooker_bubbles_seq \l__ooker_bubbles_first_tl
    \seq_pop_right:NN \l__ooker_bubbles_seq \l__ooker_bubbles_last_tl
    \bubble{bubblegreen}{sharp~corners=southeast}{white}{\l__ooker_bubbles_first_tl}\par
    \seq_map_inline:Nn \l__ooker_bubbles_seq
     {
      \bubble{bubblegreen}{sharp~corners=east}{white}{##1}\par
     }
    \bubble{bubblegreen}{sharp~corners=northeast}{white}{\l__ooker_bubbles_last_tl}\par
   }
 }
\NewEnviron{leftbubbles}
 {
  \raggedright\sffamily
  \seq_set_split:NnV \l__ooker_bubbles_seq { \par } \BODY
  \int_compare:nTF { \seq_count:N \l__ooker_bubbles_seq < 2 }
   {
    \bubble{bubblegray}{rounded~corners}{black}{\BODY}
   }
   {
    \seq_pop_left:NN \l__ooker_bubbles_seq \l__ooker_bubbles_first_tl
    \seq_pop_right:NN \l__ooker_bubbles_seq \l__ooker_bubbles_last_tl
    \bubble{bubblegray}{sharp~corners=southwest}{black}{\l__ooker_bubbles_first_tl}\par
    \seq_map_inline:Nn \l__ooker_bubbles_seq
     {
      \bubble{bubblegray}{sharp~corners=west}{black}{##1}\par
     }
    \bubble{bubblegray}{sharp~corners=northwest}{black}{\l__ooker_bubbles_last_tl}\par
   }
 }
\ExplSyntaxOff

\title{Open-vocabulary Mobile Manipulation in Unseen Dynamic Environments with 3D Semantic Maps}

\author{
    Dicong Qiu$^{1\diamond}$, Wenzong Ma$^{1\diamond}$, Zhenfu Pan$^2$, Hui Xiong$^{1\ast}$, Junwei Liang$^{1\ast}$ \\
    $^1$Hong Kong University of Science and Technology (Guangzhou) \quad $^2$Jacobi.ai 
}

\begin{document}
\maketitle

\let\thefootnote\relax\footnotetext{${\diamond}$ Equal contribution.}
\let\thefootnote\relax\footnotetext{${\ast}$ Corresponding authors.} 

\begin{abstract}
    Open-Vocabulary Mobile Manipulation (OVMM) is a crucial capability for autonomous robots, especially when faced with the challenges posed by unknown and dynamic environments.
    This task requires robots to explore and build a semantic understanding of their surroundings, generate feasible plans to achieve manipulation goals, adapt to environmental changes, and comprehend natural language instructions from humans.
    To address these challenges, we propose a novel framework that leverages the zero-shot detection and grounded recognition capabilities of pretraining visual-language models (VLMs) combined with dense 3D entity reconstruction to build 3D semantic maps.
    Additionally, we utilize large language models (LLMs) for spatial region abstraction and online planning, incorporating human instructions and spatial semantic context.
    We have built a 10-DoF mobile manipulation robotic platform \textit{JSR-1} and demonstrated in real-world robot experiments that our proposed framework can effectively capture spatial semantics and process natural language user instructions for zero-shot OVMM tasks under dynamic environment settings, with an overall navigation and task success rate of 80.95\% and 73.33\% over 105 episodes, and better SFT and SPL by 157.18\% and 19.53\% respectively compared to the baseline.
    Furthermore, the framework is capable of replanning towards the next most probable candidate location based on the spatial semantic context derived from the 3D semantic map when initial plans fail, keeping an average success rate of 76.67\%.
\end{abstract}



\section{Introduction}

Mobile manipulation is a vital and fundamental capability of autonomous robots. The recent surge of pretraining LLMs and VLMs, 
along with their integration with robotics, has drawn significant attention in research, particularly in the areas of open-vocabulary and zero-shot capabilities for autonomous robots in navigation and mobile manipulation tasks~\cite{yenamandra2023ovmm, gervet2023navigating, brohan2022rt, stone2023openworld, brohan2023can, majumdar2022zson}.
Although recent studies have explored robot manipulation in both semantic navigation~\cite{gervet2023navigating, majumdar2022zson, chaplot2020object} and open-vocabulary \cite{huang2023voxposer, yenamandra2023ovmm, lu2023ovir, chang2023context} settings, they often assume either a static environment \cite{huang2023visual, majumdar2022zson} or a non-mobile robot fixed on a tabletop \cite{huang2023voxposer, zitkovich2023rt}, and sometimes operate purely in simulation \cite{majumdar2022zson, al2022zero}.
These settings limit the capability of putting a moving robotic platform into real-world use.
Additionally, the lack of prior knowledge about an unseen environment and the dynamic factors leading to changes in the setup further complicate the problem. 
However, addressing these two problems is crucial for developing robots to become generalists and be practically applicable to a wider spectrum of real-world tasks.

To address the above challenges, we propose a novel two-stage framework enabling robots to explore and build up semantic understanding of an unseen open environment, generate feasible efficient plans by taking environment semantic context into consideration, overcoming dynamic changes of the environment, and understand human instructions and hints in natural language. At the 3D semantic mapping stage, a robot explores the environment with heuristic algorithms \cite{yamauchi1997frontier, sun2020frontier, cao2022autonomous} where sequential observations from the robot goes into a simultaneously location and mapping (SLAM) pipeline \cite{campos2021orb, campos2020inertial, mur2017orb, mur2015orb} to reconstruct dense 3D structure of the environment for navigation, and a semantics extraction and abstraction pipeline leveraging the open-vocabulary detection and zero-shot abstract reasoning capabilities of LLMs \cite{achiam2023gpt} and VLMs \cite{ren2024grounded, zhang2023recognize, chen2023internvl} to build up semantic understanding of the environment captured in a 3-layer 3D semantic map (3DSMap) for open-vocabulary navigation and mobile manipulation. At the semantics-aware open-vocabulary mobile manipulation stage, the robot parses human instructions and hints given in natural language, and comes up with corresponding semantically optimal region search plans with LLMs. When it finds the target object to fetch using the open-vocabulary detection (OVD) capability of VLMs \cite{ren2024grounded, chen2023internvl}, traditional search-based and probabilistic path and motion planners \cite{sucan2012ompl, kuffner2000rrtconnect, hart1968formal, fox1997dynamic} can take over to pick up the target object and return it to the user. Compared to traditional learning-based frameworks \cite{zitkovich2023rt, brohan2022rt, majumdar2022zson, anderson2018vision} that usually require intensive training, inspired by \cite{huang2023visual, lu2023ovir, huang2023voxposer, rana2023sayplan}, the approach we propose does not requires training by taking the most advantages of pretrained foundation models to understand and reason about the environment semantics and open-vocabulary region and instance concepts in zero-shot.

\begin{figure}[ht]
    \centering
    \includegraphics[width=1.\linewidth]{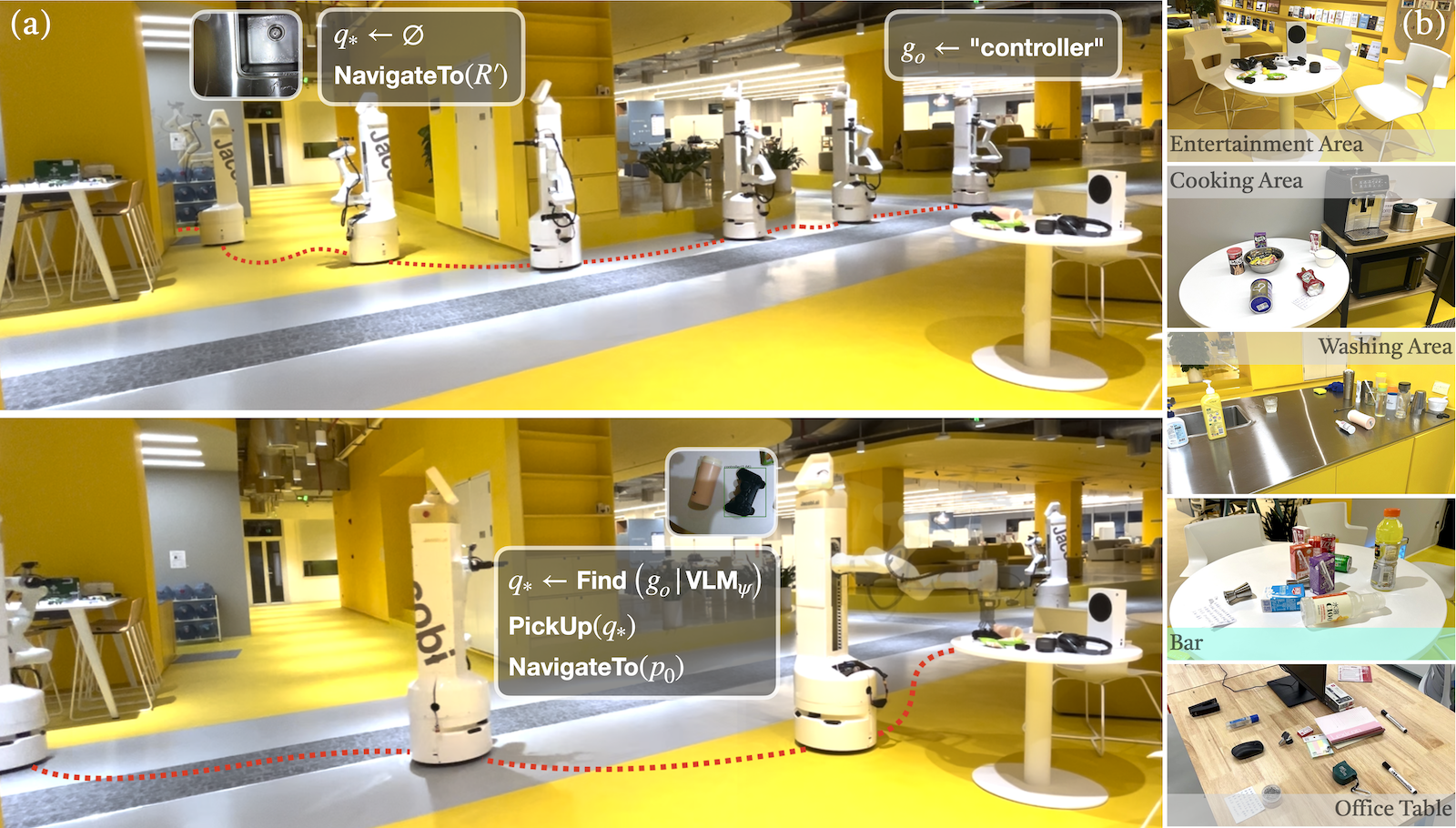}
    \caption{An illustration of our real-world experiment on OVMM with the \textit{JSR-1} robot platform we built.
    (a) A sample run in our experiment. The robot receives a misleading instruction "\texttt{fetch the controller from the washing area}" from the user. Obeying the instruction given by the user, it prioritizes the \texttt{washing area} as the first region to search, where it fails to to find the target object $g_{o} = \texttt{"controller"}$. $ R' = \texttt{entertainment area} $ follows as the next most relevant semantic region, and \textit{JSR-1} successfully finds $g_{o}$, picks up the correct corresponding instance $q^{*}$, and return to $p_{0}$.
    (b) The 5 regions in our experiment scene setup 
    with 20 different categories of objects scattered within them conforming commonsense in daily life.
    }
    \label{fig:experiment}
\end{figure}

We validate the effectiveness and robustness of the proposed framework on the mobile manipulation robotic platform \textit{JSR-1} we built. Our experiment (Sec. \ref{sec:experiment}) shows the capability of the approach we proposed in taking in natural language human instructions for the zero-shot open-vocabulary mobile manipulation task in dynamic environments, and replanning towards the next most possible location in accord to the spatial semantic context from the 3D semantic map, if the target object does not present at the location given by the prior knowledge from the user.

The main contributions of this work includes 
(i) a novel two-stage approach for a robot to explore and build up semantic understanding of an unseen open environment a zero-shot and efficiently tackle mobile manipulation tasks in the real world in open-vocabulary and dynamic environment settings, 
(ii) a 3-layer structure of 3D semantic map representation to capture not only the structural information but also the instance and abstract region semantics of the environment, (iii) a proposal-approval work flow with generic-proposed VLMs to effectively reduce the false positive rate of OVD models, and 
(iv) demonstrating the effectiveness of pretrained LLMs and VLMs on complex real-world robotic tasks with dynamic factors and semantic commonsense of an environment being considered, dispensing with the need of intensive training or fine-tuning neural network parameters.


\section{Related Work}

\textbf{Pretraining foundation models for robotics.} 
Pretraining foundation models \cite{zhou2023comprehensive, wu2023brief, li2024multimodal}, especially pretraining LLMs \cite{touvron2023llama, ye2023comprehensive, achiam2023gpt, team2023gemini, zheng2024judging} and VLMs \cite{chen2023internvl, liu2023llava, li2022languagedriven, zhou2022detecting, li2021grounded, zhang2022glipv2, radford2021learning}, have demonstrated remarkable zero-shot capabilities in a wide range of tasks \cite{liu2024llavanext, zhang2024vision, yang2024depthanything, kirillov2023segany, cheng2023segment, liu2023grounding}. 
Leveraging such advantages in robotic and embodied applications has been under active research in recent years.
\cite{zeng2022socratic} and \cite{shah2023lm} combine multiple pretrained models as submodules for visual-language manipulation and navigation. \cite{huang2023visual} directly fuses pretrained CLIP features \cite{radford2021learning, li2022languagedriven} 
for language conditioned navigation. As for planning and decision making, \cite{brohan2023can, song2023llm, huang2022language} decompose high-level tasks with LLMs into feasible plans consisting of pretrained or predefined executable actions. \cite{rana2023sayplan, liu2023llm, silver2022pddl} solve long-horizon robot planning problems with LLMs by incorporating classical planners. \cite{brohan2022rt, zitkovich2023rt, majumdar2022zson} adapt the weights of pretraining vision-language models to train end-to-end models that directly map robot observations to low-level control actions, while \cite{huang2023voxposer, liang2023code} propose training-free approaches with pretraining foundation models to achieve similar purpose. In this work, 
we utilize the zero-shot and abstract reasoning capabilities of pretraining VLMs and LLMs to address semantic-aware OVMM tasks, allowing real-world robots to perform complex tasks in unseen and dynamic environment settings without specialized training or fine-tuning, which has significant potential to generalize across various robotic platforms and tasks.

\textbf{Scene reconstruction and semantic mapping.} Reconstruction of environment facilitates robot navigation and manipulation by providing structural context of its surroundings.
Popular feature-based SLAM methods such as ORB-SLAM \cite{murartal2015orbslam, mur2017orb, campos2020inertial, campos2021orb} and VINS \cite{qin2018online, qin2018vins} support monocular and visual-inertial RGB-D SLAM with loop closure for online applications, requiring less computation than traditional offline multi-view geometric dense 3D reconstruction approaches \cite{schoenberger2016sfm, schoenberger2016mvs, schoenberger2016sfm}. Recent advances in neural radian fields \cite{mildenhall2021nerf, martin2021nerf} and Gaussian splatting \cite{kerbl20233d} have been further developed to reconstruct continuous and dense 3D scenes and instance representations for robotic applications in mapping and localization \cite{moreau2022lens, zhu2022nice, yang2022vox, matsuki2024gsslam}, navigation \cite{adamkiewicz2022vision}, object pose estimation \cite{li2023nerf} and manipulation \cite{lu2024manigaussian, dai2023graspnerf, simeonov2022neural, li2024objectaware}. Moreover, \cite{chen2023open, lu2023ovir, chang2023context} attach entity and spatial semantics captured in 3D scene graph \cite{armeni20193d} or implicit representations \cite{hsu2023ns3d}, on top of structural reconstruction, enabling spatial semantic awareness of robots in task planning, entity localization, navigation, etc. In this work, we introduce 3DSMaps, a novel 3-layer structural representation to capture both spatial structure and semantics in one, and demonstrate the effectiveness of it in our experiment.

\textbf{Mobile manipulation and navigation in the open world} requires open-vocabulary and strong adaptive capabilities of a robot. Although recent progress on pretraining foundation models have inspired more research focus, it remains an open problem \cite{yenamandra2023ovmm}. \cite{brohan2022rt, zitkovich2023rt, majumdar2022zson, chaplot2020object} propose to address this problem in end-to-end pipelines. On the other hand, \cite{rana2023sayplan, huang2023voxposer, liang2023code} introduce training-free approaches with pretraining foundation models to tackle this challenge. Prior studies often assume either a static environment \cite{huang2023visual, majumdar2022zson} or a non-mobile robot setting \cite{huang2023voxposer, zitkovich2023rt}, and sometimes operate purely in simulation \cite{majumdar2022zson, al2022zero}. We propose a novel framework in this study to holistically address the aforementioned challenges of robotic mobile manipulation in the open world with open-vocabulary, dynamic and unseen environment settings, demonstrating its potential to bring robots into real-world use.


\section{Problem Settings and Formulation}

We follow the open-vocabulary object navigation and mobile manipulation problem settings proposed by \cite{yenamandra2023ovmm, gervet2023navigating, chaplot2020object} and consider a more interactive problem setting which can be found more common in the real world. A robot is ask by a human user to find and fetch an object for the user in open-vocabulary settings. Aiming for generality and practicability of applying our framework to real-world scenarios, we do not assume any prior knowledge and further consider dynamic factors of the environment. The robot needs to explore and learn about the structural and semantic information about the environment. And at online task execution stage, objects may not remain where they were during exploration. The user may optionally provide suggestions to the robot about where the object may be located, but such suggestions can be sometimes misleading as well in reality, which is taken into consideration in our problem definition.

The goal of the above task can be formally described as a $ (g_{o}, g_{R}) $-tuple. The robot receives a natural language instruction $ \mathcal{L} $ from the user about the goal, a typical example is "\texttt{fetch something [from somewhere]}", where "\texttt{something}" suggests the target object $ g_{o} $ to fetch and "\texttt{somewhere}" is a hint from the user prior knowledge about the possible target region $ \hat{g}_{R} $ where $ g_{o} $ may be located. Note that the hint $ \hat{g}_{R} $ given by the user is optional and it can sometimes be misleading due to false memory of the user or dynamic changes of the environment. In a successful run, the robot shall reach the target region $ g_{R} $, pick up the target object $ g_{o} $ and return it to the user.

Real-world scenarios are rather complicated, to avoid ambiguity and alleviate excessive engineering, there are two reasonable assumptions. (i) In any scene, there can be multiple types of objects and sometimes multiple objects in the same type presenting in the scene or even the same location. We may assume either the target object is the only one object with the same type presenting in the scene, or if there are multiple objects with the same type as the target object, fetching any of them is considered successful. (ii) View angle planning itself, especially in 3D settings, is an active standalone research topic. To focus on the core problem to tackle, we may assume one or several predefined view angles (camera poses) of the robot at each location in a scene. A robot can navigate and pose its camera to these predefined angles in search for the target object.


\section{Method}

Prior knowledge about the environment and exact region to fetch the target object from are inaccessible to a robot. Hints from users are optional and not assumed to be reliable. Beyond the open-vocabulary settings in mobile manipulation, it poses challenges on both building up structural and semantic knowledge about the environment, and effectively and efficiently leverage such knowledge to complete the aforementioned task. To that end, we propose a two-stage framework to holistically tackle the above problem in a training-free manner. At the 3D semantic mapping stage (Sec. \ref{sec:semantic_mapping}), heuristic exploration and reconstruction from feature-based SLAM are followed by instance semantics extraction and region semantics abstraction, taking the zero-shot and abstract reasoning advantages of VLMs and LLMs for the robot to learn spatial structure and environment semantics in a training-free manner. At the semantics-aware open-vocabulary mobile manipulation stage (Sec. \ref{sec:semantic_ovmm}), the robot prioritizes the regions to search with LLMs in accord to the user instruction and the environment semantics, picks up the target object detected by open-vocabulary detection models and returns it to the user.

\begin{figure}[ht]
\begin{subfigure}{.36\textwidth}
  \centering
  \includegraphics[width=0.95\linewidth]{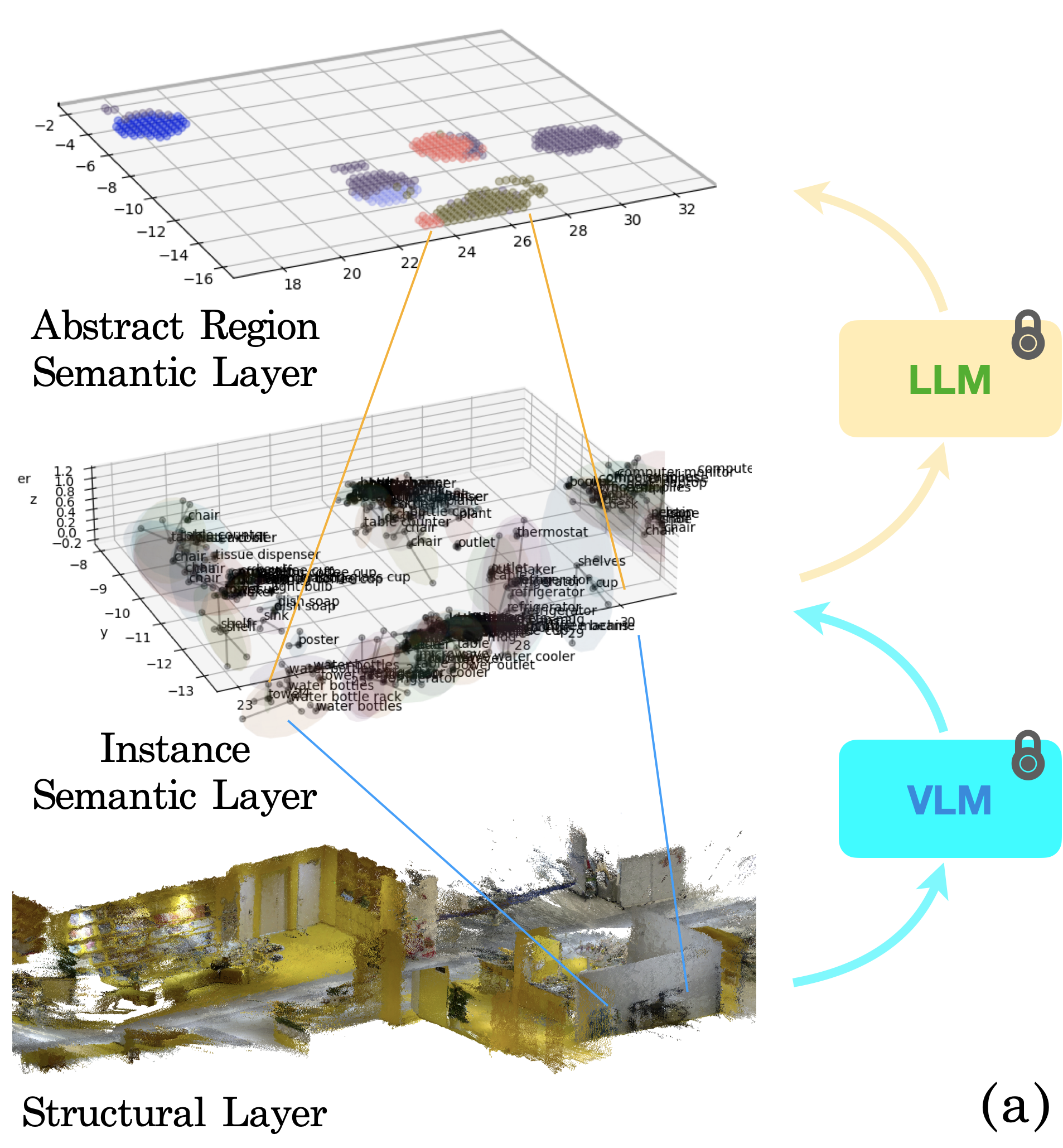}
\end{subfigure}%
\begin{subfigure}{.64\textwidth}
  \centering
  \includegraphics[width=0.95\linewidth]{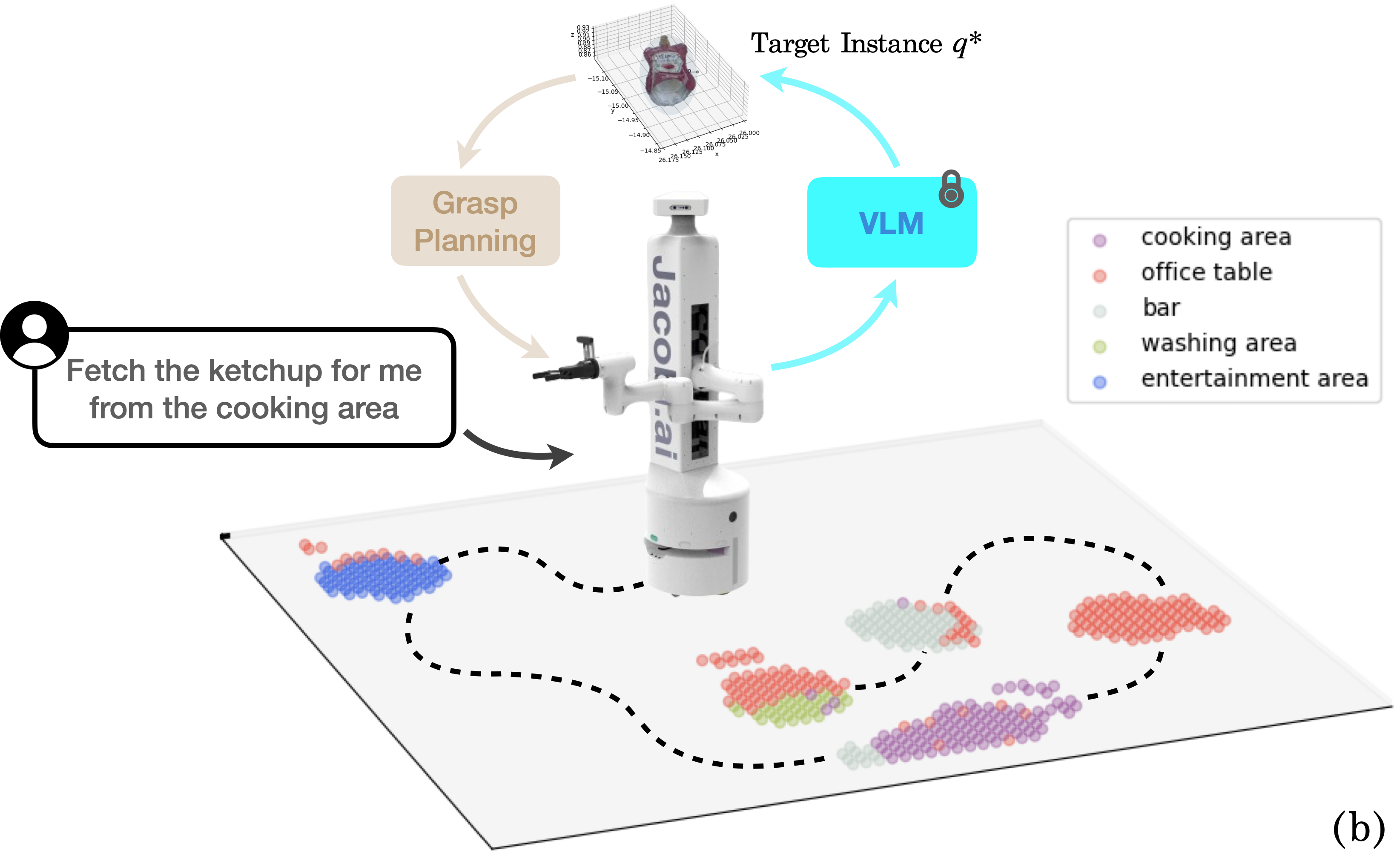}
\end{subfigure}
\caption{An overview of our proposed two-stage framework. (a) A 3-layer structured \textit{3DSMap} is built at the 3D semantic mapping stage, by leveraging heuristic exploration and feature-based SLAM to build the \textit{structural layer}, using
pre-trained VLMs and LLMs to extract instances from and propose region divisions for an environment to build the \textit{instance semantic layer} and the \textit{abstract region semantic layer}, respectively. (b) At the semantics-aware OVMM stage, a robot takes both region semantics from \textit{3DSMap} and user instructions as context to prioritize regions to search with LLMs and fetch the target object for the user.}
\label{fig:framework_overview}
\end{figure}

\subsection{3D Semantic Mapping in Unseen Environments}\label{sec:semantic_mapping}

In a completely unseen environment with no prior knowledge, it is essential for a robot to become aware of the environment structure and further build up semantic understanding of the environment. In this section, we propose a 3-layer (structural layer, instance semantics layer and abstraction region layer) structure of 3D semantic maps as illustrated in fig.\ref{fig:framework_overview}(a) to capture the both structure and semantics of the environment.

\textbf{Heuristic exploration and structural reconstruction.} Off-the-shelf heuristic frontier exploration algorithms \cite{yamauchi1997frontier, sun2020frontier, cao2022autonomous, cao2023representation} are used for the robot to explore the environment. Sequential vision-inertial sensing input from the RGB-D camera and the IMU of the robot is recorded, consisting of RGB image frames $\{ \mathcal{I}^{t}_{\text{RGB}} \}$ and depth frames $\{ \mathcal{I}^{t}_{\text{D}} \}$ along with the IMU readings $\{ I^{t} \}$, each of which is associated with a timestamp $ t $. We adopt ORB feature based RGB-D-I simultaneous localization and mapping (SLAM) algorithms \cite{campos2021orb, campos2020inertial, mur2017orb, mur2015orb} to fuse the RGB-D camera data and the IMU readings, build 3D feature maps and determine key frames from loop-closure corrected camera poses trajectory $ \{ T^{\text{cam}, k} \}_{k=1}^{K} $ with $ K $ being the total number of key frames.

The camera poses combined with the RGB and depth data  associated to key frames in the sensor data sequence generates a key frame trajectory as a sequence of $ ( \mathcal{I}^{k}_{\text{RGB}}, \mathcal{I}^{k}_{\text{D}}, T^{k}_\text{cam} ) $-tuples. Each depth key frame $ \mathcal{I}^{k}_{\text{D}} $ is by nature associated with a list of homogeneous 3D points $ \mathbf{P}^{k}_{\text{cam}} = \{ p^{(k,j)}_{\text{cam}} \}_{j} $ in the camera coordinate frame by the intrinsic parameters of the camera \cite{szeliski2022computer}. Dense reconstruction of the 3D environment structure in point cloud representation can be achieved by re-projecting the dense points from the depth frames onto the global 3D space $ \mathbf{P}^{k} = T^{\text{cam}, k} ( \mathbf{P}^{k}_{\text{cam}} )^{\mathsf{T}} $ and accumulating them \cite{mur2017orb} for localization and navigation. We can further color the point cloud by the re-projection and association of the 3D points to corresponding RGB image frames for visualization. The dense reconstruction of the environment structure composes the \textit{structural layer} of a 3DSMap.

\textbf{Geometric and semantic extraction of 3D instances.} To build semantic awareness of the environment in a training-free manner inspired by \cite{huang2023visual} and \cite{lu2023ovir}, we adopt the Grounded SAM pipeline \cite{ren2024grounded} for 2D open-vocabulary detection and pixel-wise segmentation. For each RGB image frame at key frame $ k $, it detects object instances $ \left\{ b^{(k,i)} \right\}_{i=1}^{N_{k}} $ with open-vocabulary detection models, such as Grounding DINO \cite{liu2023grounding} or Detic \cite{zhou2022detecting}, and segments the detected instances with SAM \cite{kirillov2023segany} into 2D pixel masks $ \left\{ m^{(k,i)} \right\}_{i=1}^{N_{k}} $. Text prompt inputs required by OVD models consisting of instance proposals can be automatically generated by open-set image tagging models \cite{huang2023open}, captioning models \cite{huang2023tag2text, li2022blip} or generic-purposed VLMs \cite{chen2023internvl} (Appx. \ref{appendix:vlm_for_instance_proposals}). Re-projection of the 3D points $ \mathbf{P}^{k} $ from depth frames onto each of pixel mask $ m^{(k,i)} $ further segments these 3D points into point-based geometric representations $ \left\{ \mathbf{P}^{(k,i)} \right\}_{i=1}^{N_{k}} $ of the instances presenting in the scene, each of which is decomposed into its geometric center $ \bar{p}^{(k,i)} = \frac{1}{N_{k}}\sum_{p \in \mathbf{P}^{(k,i)}} p $ and relative geometry $ \bar{\mathbf{P}}^{(k,i)} = \mathbf{P}^{(k,i)} - \bar{p}^{(k,i)} $. Associated with object instance semantics $ b^{(k,i)} $, especially the label of the instance, an instance geometric representation becomes a semantic geometric instance $ q^{(k,i)} = \left( b^{(k,i)}, \bar{p}^{(k,i)}, \bar{\mathbf{P}}^{(k,i)} \right) $ and is then registered by its spatial coordinate $ \bar{p}^{(k,i)} $ onto the \textit{instance semantics layer} of a 3DSMap. The geometry $ \mathbf{P}^{(k,i)} $ of an instance from a key frame by the nature of field of view is a partial geometric representation it, and multiple $ \mathbf{P}^{(k,i)} $'s from different key frames $ k $s may point to the same physical instance. Instance fusion \cite{lu2023ovir} across key frames can effectively resolve this problem, however, it is not strictly required by our proposed framework.

\textbf{Region semantics abstraction.} We consider robots operating in large open scenarios. In real world, a scene in general consists of multiple functional areas with different semantic context. Efficient utilization of such abstract semantics for planning is beneficial for a robot to overcome dynamic factors in an environment, i.e. objects may not always persist where they were when first observed. To this end, instances $ \{ q^{(k,i)} \} $ extracted and registered on the instance semantic layer are projected onto the 2D floor plane by simply removing the height dimension from their geometric center $ \bar{p}^{(k,i)} $, forming a 2D bird-eye-view (BEV) instance semantics map. We place a circular sliding window with radius $ r $ on the BEV map, starting from the top-left corner and swiping through the entire map by certain step size $ \Delta d $. At step $(s_{x}, s_{y})$, the area within the sliding window $ \left\{ (x,y) | (x - s_{x}\Delta d)^2 + (y - s_{y}\Delta d)^2 \leq r^{2} \right\} $ selects a set of object instances falling inside, whose labels after repeated terms removed effectively describe the objects presenting within this area. We then leverage the zero-shot abstract reasoning capability of LLMs \cite{achiam2023gpt} to come up with a list of region proposals, depicting the abstract region semantics of the area (Appx. \ref{appendix:llm_for_region_abstraction}). After a full sweep, a zero-shot dense prediction of region semantics $ R(x,y) $ over the entire BEV space is therefore generated, constituting the \textit{abstract region layer} of a 3DSMap, providing abstract semantic information about different regions in an environment. 

\subsection{Semantics-aware Open-vocabulary Mobile Manipulation}\label{sec:semantic_ovmm}

With structural and semantic knowledge built up about the unseen environment and captured in 3DSMaps, a robot can efficiently leverage such knowledge to complete OVMM tasks and withstand dynamic changes in the environment conforming semantic commonsense. Alg.\ref{alg:semantic_ovmm} depicts the overall procedure of open-vocabulary mobile manipulation considering environment semantic context.

\begin{algorithm}[ht]
\caption{Semantics-aware open-vocabulary mobile manipulation}\label{alg:semantic_ovmm}
\begin{algorithmic}
\Require natural language user instruction $ \mathcal{L} $, random sequence of regions in the scene $ \mathcal{S}_{R} $
\Ensure robot starts nearby the user at $p_{0}$
\State $(g_{o}, \hat{g}_{R}) \gets \textbf{LLM}_{\theta}(\mathcal{L} | \mathcal{T}_{\text{Parsing}})$  \Comment{parse user instruction}
\State $ \mathcal{S}_{R}^{*} \gets \textbf{LLM}_{\theta} (\mathcal{S}_{R}, g_{o} | \mathcal{T}_{\text{Prioritization}})$  \Comment{prioritize regions by semantic relevance with LLM}
\If { $ \hat{g}_{R} \neq \emptyset $ }
    \State $ \mathcal{S}_{R}^{*} \gets \text{Reprioritize} (\mathcal{S}_{R}^{*}, \hat{g}_{R})$  \Comment{assign highest priority to $\hat{g}_{R}$}
\EndIf
\For { $ R \in \mathcal{S}_{R}^{*} $ }
    \For { $ p \in \text{Searchable}(R) $ }
        \State $ \text{NavigateTo}(p) $  \Comment{navigate to a searchable location $p$ in region $R$}
        \State $ q \gets \text{Find}(g_{o} | \textbf{VLM}_{\psi}) $  \Comment{attempt to find $g_{o}$ leveraging VLM}
        \If { $ q \neq \emptyset $ }
            \If { $ \text{PickUp}(q) $ }  \Comment{attempt to pick up instance $q$}
                \State $ \text{NavigateTo}(p_{0}) $  \Comment{bring it back to the user}
            \EndIf
        \EndIf
    \EndFor
\EndFor
\end{algorithmic}
\end{algorithm}

\textbf{Open-vocabulary semantic prioritization for search regions.} The robot receives a natural language instruction $ \mathcal{L} $ from the user about the goal of the mobile manipulation task, asking the robot the fetch a target object $g_{o}$. The user may optionally provide hints about at which region $\hat{g}_{R}$ the target object may be. We parse $\mathcal{L}$ and extract $g_{o}$ and $\hat{g}_{R}$ from it, using pretrained LLMs \cite{achiam2023gpt} with prompt template for instruction parsing $\mathcal{T}_{\text{Parsing}}$ (Appx. \ref{appendix:llm_for_instruction_parsing}). The target object $g_{o}$ along with a list of regions $\mathcal{S}_{R}$ presenting in the scene obtained from region semantics abstraction (Sec. \ref{sec:semantic_mapping}) are input into the pretrained LLMs with template $\mathcal{T}_{\text{Prioritization}}$ to prioritize the regions by semantic relevance between the target object $ g_{o} $ and each of the region in $\mathcal{S}_{R}$ (Appx. \ref{appendix:llm_for_regions_prioritization}), which is followed by an optional re-prioritization step to assign highest priority to the region $\hat{g}_{R}$ suggested by the user in $\mathcal{L}$. The final outcome from the prioritization step (and re-prioritization step) is an ordered sequence of regions $S_{R}^{*}$ indicating the search priority of different regions for $g_{o}$.

\textbf{Prioritized navigation and in-region exploration.} Following the prioritized list $S_{R}^{*}$ of regions to search, the robot attempts to reach these regions one after another and find the target object $g_{o}$. Since we consider mobile manipulation of grounded robots, though in general our framework is applicable to any 3D spatial robots, to reduce computational complexity, the structural layers of 3DSMaps is flattened into 2D cost maps \cite{matthies1988integration, moravec1988sensor} for navigation. Heuristic reachability analysis \cite{vahrenkamp2013robot, makhal2018reuleaux} is adapted to pre-compute a list 2D searchable locations within a region with overlapped and infeasible locations filtered out. Global navigation trajectories are planned using a search-based global planner \cite{hart1968formal, sucan2012ompl} and followed by the DWA local planner \cite{gerkey2008planning, fox1997dynamic}, towards each of the searchable locations $ p \in \text{Searchable}(R) $ within a region $R$ then another following the prioritized list $S_{R}^{*}$. At each searchable location $p$, the robot attempts to find and pick up the target object $g_{o}$, whose details will be further discussed below. At failure of finding $g_{o}$ at $p$, the robot will head to the next searchable location and repeat the above procedure as illustrated in alg. \ref{alg:semantic_ovmm}, allowing the robot to complete the task efficiently following the optimal semantic search path.

\textbf{Open-vocabulary instance detection and manipulation.} Reaching at a searchable location $p$ in region $R$, the robot will then adopt \cite{sucan2012ompl, kuffner2000rrtconnect} to conduct end-effector planning for a camera pose $T^{\text{cam}}$ looking towards (e.g. downwards on) the operation area at $p$. Similar pipeline \cite{ren2024grounded, liu2023grounding, kirillov2023segany} from Sec. \ref{sec:semantic_mapping} is reused for OVD and pixel-wise segmentation, and the prompt instructing the OVD models is simply tha target object $g_{o}$. However, OVD models suffer from high false-positive rate \cite{pham2024lp}, significantly reducing the overall reliability. Therefore, we propose a proposal-approval work flow, with an OVD model \cite{ren2024grounded} coming up with detection proposal and another VLM model \cite{chen2023internvl} double checking the result to either approve or reject. The robot then plans the end-effector trajectory of its gripper for grasping \cite{kuffner2000rrtconnect, sucan2012ompl}, towards the semantic geometric instance $ q^{*} = \left( b^{*}, \bar{p}^{*}, \bar{\mathbf{P}}^{*} \right) $ with the highest confidence from the approved list of detected instances. We allows at most $ n_{e} = 3 $ trials on grasping at each location $p$. After success in picking up $q^{*}$, the robot will return to the user.


\section{Experiment}
\label{sec:experiment}

We analyze the effectiveness and performance of our proposed method in the a large real-world open space
with the \textit{JSR-1} mobile manipulation robotic platform we built. Our experiment covers 135 independent episodes (eps.) in total with real robot for quantitative evaluation, which are split into 5 experiment groups. Experiment details are presented and analyzed in below.

\textbf{Experiment settings.}
Our robotic experiment is conducted in a large real-world indoor open space covering an area of over 200 $m^2$. Within it, we set up 5 regions as shown in fig.\ref{fig:experiment}(b) along with 20 different categories of objects placed within these regions conforming daily commonsense 
as shown in tab.\ref{tab:exp_groups_main}.
Our quantitative experiment with real robot platform consists of 135 independent episodes in total, which are divided into 5 experiment groups (Appx.\ref{appendix:experiment_setup}). At the beginning of each run, the robot starts from the same position $p_{0}$ nearby the user and receives a natural language instruction.

\begin{table}[h!]
\begin{center}
\caption{Division of Experiment Groups}
\label{tab:exp_groups_main}
\begin{tabular}{c|c|l}
\hline
Exp. Group & \# of Eps. & Description \\
\hline
NoHint & 45 & User instructions do not contain region hints \\
Random & 30 & Control group, where region search orders are randomized \\
Hinting & 30 & User instructions contain region hints \\
ErrantSemantics & 15 & Target objects are placed in semantic irrelevant regions \\
Misleading & 15 & User instructions contain misleading region hints \\
\hline
\end{tabular}
\end{center}
\end{table}

\textbf{Robot platform.} 
We have built a 10-DoF mobile manipulation semi-humanoid robotic platform \textit{JSR-1}, the robot appeared in fig.\ref{fig:experiment}(a), which consists of a 2-DoF wheeled chassis, a 6-DoF robotic arm with a 1-DoF gripper, and an RGBD camera on nearby its end-effector. Besides, a 1-DoF waist link assembles its chassis and arm, extending its operation range from 0 to 200cm in height.

\textbf{Evaluation metrics.} 
We introduce 5 metrics for quantitative evaluation. 
Success on first trial (\textbf{SFT}) is the ratio of episodes where the first semantic proposal for search region contains the target object. 
Success on navigation (\textbf{SN}) is the ratio of episodes where the robot has navigated to nearby the target object.
Success on picking (\textbf{SP}) is the ratio of episodes where the robot has picked up the target object successfully.
Overall success rate (\textbf{Succ.}) is the ratio of episodes where the robot has successfully carried the target object to $p_{0}$.
Success weighted by path length $\textbf{SPL} = \frac{1}{N} \sum_{i=1}^{N} S_i \frac{ l_i }{ \max (p_i, l_i) }$ measures the efficiency of reaching the goal in addition to the success rate \cite{anderson2018evaluation}.

\begin{table}[ht]
\begin{center}
\caption{The overall performance of proposed method evaluated through real-world robot experiment. We compare the result of each experiment groups with the control group (Random), and report the SFT increased (SFT Incr.) and SPL increased (SPL Incr.) for each group respectively.}
\label{tab:overall_performance_result}
\begin{tabular}{c|c c c |c c|c c}
\hline
Group. Metrics & SN & SP & Succ. & SFT & SPL & SFT Incr. & SPL Incr. \\
\hline
Random (R.) & 26/30 & 23/30 & 23/30 & 7/30 & 0.5332 & - & - \\
\hline
NoHint & 34/45 & 33/45 & 33/45 & 33/45 & 0.6878 & \textbf{214.31\%} & \textbf{31.46\%}\\
Hinting & 25/30 & 21/30 & 21/30 & 30/30 & 0.6841 & \textbf{328.63\%} & \textbf{30.75\%}\\
ErrantSemantics & 13/15 & 13/15 & 13/15 & 0/15 & 0.4942 & -100.00\% &-5.54\%\\
Misleading & 13/15 & 10/15 & 10/15 & 0/15 & 0.4519 & -100.00\% & -13.63\%\\
\hline
Total without R. & 80.95\% & 73.33\% & 73.3\% & 60.00\% & 0.6254 & \textbf{157.18\%} & \textbf{19.53\%} \\
\hline
\end{tabular}
\end{center}
\end{table}

\textbf{Experiment Result Analysis.}
By the experiment result (tab.\ref{tab:overall_performance_result}), 
our proposed method demonstrates a decent performance and robustness in complex real-world OVMM tasks,
achieving an overall success rate of 73.33\% and 
a successful navigation rate of 
80.95\%, under various situations, including objects being randomly placed in semantic irrelevant regions and user giving misleading instructions. Compared to the control group (Random), it has better overall performance on SFT and SPL by 157.18\% and 19.53\% respectively. As for normal situations without misplacement of objects or misleading user instructions, our approach demonstrates a significant performance advantage in the NoHint and Hinting groups, with better SFT by 214.31\% and 328.63\%, and SPL by 31.46\% and 30.75\%. The result shows that our proposed method is able to efficiently incorporate spatial region semantics and user hints for semantic-aware OVMM tasks, and it can robustly recover from failure and complete the tasks even being exposed to dynamic factors and misleading instructions.

\begin{table}[ht]
\begin{center}
\caption{Comparison of the experiment results between NoHint and Hinting groups.}
\label{tab:exp_result_nohint_vs_hinting}
\begin{tabular}{c|c c c c c}
\hline
Exp. Group & SFT (\%) & SN (\%) & SP (\%) & Succ. (\%) & SPL (0.0-1.0) \\
\hline
NoHint & 73.33 & 75.56 & 73.33 & 73.33 & 0.6878 \\
Hinting & \textbf{100.00} & 83.33 & 70.00 & 70.00 & 0.6841 \\
\hline
\end{tabular}
\end{center}
\end{table}

Furthermore, as shown in tab.\ref{tab:exp_result_nohint_vs_hinting}, the Hinting group achieves 100.00\% SFT for its leverage of region hints in the instruction from the user, indicating the effectiveness of our framework to incorporate prior knowledge and suggestions from humans. We also notice from tab.\ref{tab:overall_performance_result} that in Misleading group SPL is below average and less than the control group 13.63\%. It indicates that our framework is sensitive to human instruction, and misleading or wrong suggests can lead to lower efficiency. However, it keeps an $\textbf{SN}=13/15=86.67\%$ above the average and a reasonable overall success rate at 66.67\%, showing the failure recovering capability of our proposed framework.


\section{Conclusion and Future Work}
\label{sec:conclusion}

In this work, we propose a novel framework that tackles the problem of Open-Vocabulary Mobile Manipulation, which leverages the zero-shot detection and grounded recognition capabilities of pretraining visual-language models (VLMs) combined with dense 3D entity reconstruction to build 3D semantic maps.
Additionally, we utilize large language models (LLMs) for spatial region abstraction and online planning, incorporating human instructions and spatial semantic context.
We have built a 10-DoF mobile manipulation robotic platform \textit{JSR-1} and conducted real-world experiments to demonstrate the effectiveness of our proposed training-free method.
In future work, we will focus on incorporating autonomous exploration techniques to extend our system's capabilities to unknown environments. Furthermore, exploring the use of multiple agents or robots for collaborative exploration and scanning of environments will improve efficiency and coverage in unknown or large areas.


\clearpage

\acknowledgments{The authors would like to thank Lei Zhu from Jacobi.ai for his help during the experiment, and HKUST(GZ) MakerSpace for test site support.}


\bibliography{refs}  

\clearpage

\appendix

\section{Experiment Details of LLMs and VLMs on Zero-shot Tasks}

\subsection{VLMs for Instance Proposals for OVD Prompts}\label{appendix:vlm_for_instance_proposals}

\subsubsection{InternVL 1.5}

Below is an example of using InternVL 1.5 \cite{chen2023internvl} for the zero-shot instance proposals for OVD prompts.

\begin{figure}[ht]
    \centering
    \includegraphics[width=0.5\textwidth]{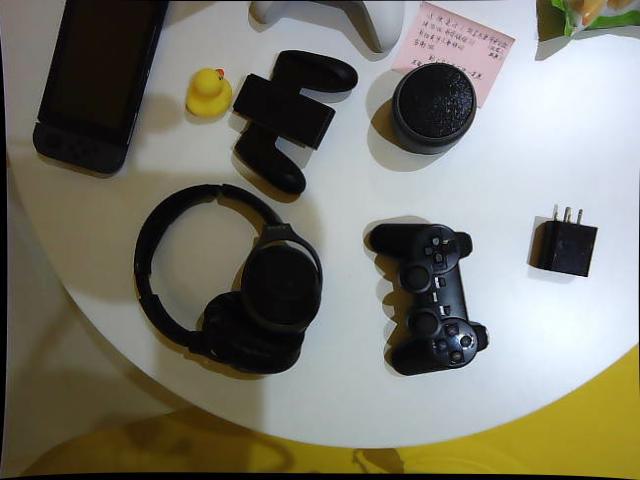}
    \label{fig:vlm_instance_proposals}
\end{figure}

\begin{rightbubbles}
    What objects are in the image? Return in JSON format \texttt{ \{"objects": object\_name, ...\} }.
\end{rightbubbles}

\begin{leftbubbles}
    \texttt{ \{"objects": ["headphones", "controller", "phone", "rubber duck", "charger", "speaker", "note", "plug"]\} }
\end{leftbubbles}

\subsection{LLMs for Zero-shot Region Abstraction Proposal}\label{appendix:llm_for_region_abstraction}

\subsubsection{GPT-4o}

Below is an example of using GPT-4o \cite{achiam2023gpt} for the zero-shot region abstraction proposal task.

\begin{rightbubbles}
    \textbf{SYSTEM:} The user will give you a list of objects inside a region and a list of region candidates in JSON format \texttt{ \{"objects": [object\_1, object\_2, ...], "region\_candidates": [region\_candidate\_1, region\_candidate\_2, ...]\} }, please order these regions in decreasing order of likelihood and return just in JSON format \texttt{ \{"region\_proposals": [region\_1, region\_2, ...]\} }, do not reply markdown format.
\end{rightbubbles}

\begin{rightbubbles}
    \textbf{USER:} \texttt{ \{ "objects": [ "beverage", "bottle", "can", "soda", "table", "juice", "mustard", "ketchup" ], "region\_candidates": [ "living room", "classroom", "coffee shop", "kitchen", "hallway" ] \} }
\end{rightbubbles}

\begin{leftbubbles}
    \textbf{ASSISTANT:} \texttt{
    \{'region\_proposals': ['kitchen', 'coffee shop', 'living room', 'classroom', 'hallway']\}
    }
\end{leftbubbles}

\subsubsection{InternVL 1.5}

Below is an example of using InternVL 1.5 \cite{chen2023internvl} for the zero-shot region abstraction proposal task.

\begin{rightbubbles}
    I will give you a list of objects inside a region, and a list of region candidates in JSON format, please order these regions in decreasing order of likelihood and return just in JSON format only: \\
    \texttt{ \{ "region\_proposals": [ region\_1, region\_2, ... ] \} }, \\
    with no verbose information or justification. Now, let’s begin. \\
    \texttt{ \{ "objects": [ "beverage", "bottle", "can", "soda", "table", "juice", "mustard", "ketchup" ], "region\_candidates": [ "living room", "classroom", "coffee shop", "kitchen", "hallway" ] \} }
\end{rightbubbles}

\begin{leftbubbles}
    \texttt{ \{
    "region\_proposals": [
    "coffee shop",
    "kitchen",
    "classroom",
    "living room",
    "hallway"
    ]
    \} }
\end{leftbubbles}

\subsubsection{LLaMA 3}

Below is an example of using LLaMA 3 \cite{touvron2023llama} for the zero-shot region abstraction proposal task.

\begin{rightbubbles}
    I will give you a list of objects inside a region, and a list of region candidates in JSON format, please order these regions in decreasing order of likelihood and return just in JSON format only: \\
    \texttt{ \{ "region\_proposals": [ region\_1, region\_2, ... ] \} }, \\
    with no verbose information or justification. Now, let’s begin. \\
    \texttt{ \{ "objects": [ "beverage", "bottle", "can", "soda", "table", "juice", "mustard", "ketchup" ], "region\_candidates": [ "living room", "classroom", "coffee shop", "kitchen", "hallway" ] \} }
\end{rightbubbles}

\begin{leftbubbles}
    \texttt{ \{ "region\_proposals": [ "kitchen", "coffee shop", "living room", "hallway", "classroom" ] \} }
\end{leftbubbles}

\subsection{LLMs for User Natural Language Instruction Parsing}\label{appendix:llm_for_instruction_parsing}

\subsubsection{GPT-4o}

Below is an example of using GPT-4o \cite{achiam2023gpt} to parse user instruction in natural language, and convert into structural instruction.

\begin{rightbubbles}
    \textbf{SYSTEM:} The user will give you an instruction in natural language about something ("target object") he/she wants to find, and the user may or may not give further guess about what region the target object may be located. Please turn the instruction into JSON format \texttt{ \{"target\_object": target\_object, "region": region\} }, where \texttt{ region } shall be set as \texttt{ null } if the user does not give further guess about region, do not reply markdown format.
\end{rightbubbles}

\begin{rightbubbles}
    \textbf{USER:} Fetch the spray cleaner from the entertainment area.
\end{rightbubbles}

\begin{leftbubbles}
    \textbf{ASSISTANT:} \texttt{
    \{'target\_object': 'spray cleaner', 'region': 'entertainment area'\}
    }
\end{leftbubbles}

\begin{rightbubbles}
    \textbf{USER:} Fetch the milk powder.
\end{rightbubbles}

\begin{leftbubbles}
    \textbf{ASSISTANT:} \texttt{
    \{'target\_object': 'milk powder', 'region': None\}
    }
\end{leftbubbles}

\subsection{LLMs for Search Regions Prioritization}\label{appendix:llm_for_regions_prioritization}

\subsubsection{GPT-4o}

Below is an example of using GPT-4o \cite{achiam2023gpt} to prioritize regions to search for the target object. In this part, we only consider the mapping from target object to a prioritized list of regions to search, without considering the region suggestion from user instruction.

\begin{rightbubbles}
    \textbf{SYSTEM:} The user will give you a list of region names in JSON format \texttt{ \{"regions": [region\_1, region\_2, ...], "target\_object": object\_name\} }, and the name of a target object he/she wants to find, please proposal a list containing the names of these regions in descending order of priority to search, and return in JSON format \texttt{ \{"ordered\_regions": [ordered\_region\_1, ordered\_region\_2, ...]\} }, do not reply markdown format.
\end{rightbubbles}

\begin{rightbubbles}
    \textbf{USER:} \texttt{\{'regions': ['entertainment area', 'washing area', 'kitchen', 'bar', 'office table'], 'target\_object': 'controller' \} }
\end{rightbubbles}

\begin{leftbubbles}
    \textbf{ASSISTANT:} \texttt{
    \{'ordered\_regions': ['entertainment area', 'bar', 'office table', 'kitchen', 'washing area']\}
    }
\end{leftbubbles}

\section{List of Symbols}

\begin{center}
\begin{tabular}{ c l } 
\hline
Symbol & Description \\
\hline
$ \mathcal{L} $ & User instruction in natural language \\ 
$ g_{o} $ & Target object to fetch \\
$ g_{R} $ & Target region where the target object is located \\
$ \mathcal{I}_{\text{RGB}} $ & Image data from RGB-D camera \\ 
$ \mathcal{I}_{\text{D}} $ & Depth data from RGB-D camera \\ 
$ T^{\text{cam}} $ & Homogeneous transformation indicating the global pose of the RGB-D camera \\ 
$ t $ & Time stamp of a timed sequence, $ 0 \leq t \leq T $ \\
$ T $ & Maximum time stamp of a timed sequence \\
$ k $ & Key frame index, $ k \in \{ 1, 2, \cdots, K \} $ \\
$ K $ & Total number of key frames \\
$ p^{(k,j)}_{\text{cam}} $ & The $j$-th homogeneous 3D point in camera frame from the $k$-th key frame \\
$ \mathbf{P}^{k}_{\text{cam}} $ & Matrix containing all 3D points in camera frame as columns from the $k$-th key frame \\
$ \mathbf{P}^{k} $ & Matrix containing all 3D points in global frame as columns from the $k$-th key frame \\
$ b^{(k,i)} $ & The $i$-th object instance detected at the $k$-th key frame with label and bounding box \\
$ m^{(k,i)} $ & Pixel mask for the $i$-th object instance detected at the $k$-th key frame \\
$ N_{k} $ & Total number of object instances detected at the $k$-th key frame \\
$ \mathbf{P}^{(k,i)} $ & Matrix containing all 3D points associated with the $i$-th instance at the $k$-th key frame \\
$ \bar{p}^{(k,i)} $ & Geometric center of the $i$-th instance at the $k$-th key frame \\
$ \bar{\mathbf{P}}^{(k,i)} $ & Relative geometry of the $i$-th instance at the $k$-th key frame \\
$ q^{(k,i)} $ & The $i$-th semantic geometric instance extracted at the $k$-th key frame \\
$ r $ & Radius of the circular sliding window for region semantics abstraction \\
$ \Delta d $ & Step size distance by which the sliding window moves for region semantics abstraction \\
$ s_{x} $ & Sliding window swiping step along the $x$-axis \\
$ s_{y} $ & Sliding window swiping step along the $y$-axis \\
$ R(x,y) $ & Region semantics suggesting a label of the region containing coordinate $ (x,y) $ \\
$ \text{Searchable}(R) $ & A list of searchable locations in region $R$ \\
$ q^{*} $ & Semantic geometric instance with the highest confidence to pick up \\
$ n_{e} $ & Maximum number of trails for grasping at a location \\
\hline
\end{tabular}
\end{center}

\clearpage

\section{Experiment Details}\label{appendix:experiment_details}

\subsection{Experiment Setup}\label{appendix:experiment_setup}
\label{sec:experiment setup}

\begin{table}[h!]
 \begin{center}
\caption{Default object placement in different regions}
\begin{tabular}{c|l}
\hline
Region & Default Objects \\
\hline
Entertainment Area & controller, toy, charger, snacks \\ 
Washing Area & sponge, cloth, cup, spray cleaner \\
Cooking Area & ketchup, milk powder \\
Bar & bottled water, cup, milk, soda \\
Office Table & marker, stapler, pen, tape, mouse, bottle glue \\
\hline
\end{tabular}
 \end{center}
\end{table}

\begin{table}[h!]
\begin{center}
\caption{Division of Experiment Groups}
\label{tab:exp_groups}
\begin{tabular}{c|c|l}
\hline
Exp. Group & \# of Eps. & Description \\
\hline
NoHint & 45 & User instructions do not contain region hints \\
Random & 30 & Control group, where region search orders are randomized \\
Hinting & 30 & User instructions contain region hints \\
ErrantSemantics & 15 & Target objects are placed in semantic irrelevant regions \\
Misleading & 15 & User instructions contain misleading region hints \\
\hline
\end{tabular}
\end{center}
\end{table}

\subsection{Experiment Result}\label{appendix:experiment_result}
.
\begin{table}[h!]
 \begin{center}
\caption{Detailed result of our robot experiment}
\begin{tabular}{c|c c c c c}
\hline
Exp. Group & SFT & SN & SP & Succ. & SPL \\
\hline
NoHint & 33/45 (73.33\%) & 34/45 (75.56\%) & 33/45 (73.33\%) & 33/45 (73.33\%) & 0.6878 \\
Random & 7/30 (23.34\%) & 26/30 (86.67\%) & 23/30 (76.67\%) & 23/30 (76.67\%) & 0.5332 \\
Hinting & 30/30 (100.00\%) & 25/30 (83.33\%) & 21/30 (70.00\%) & 21/30 (70.00\%) & 0.6841 \\
ErrantSemantics & 0/15 (0.00\%) & 13/15 (86.67\%) & 13/15 (86.67\%) & 13/15 (86.67\%) & 0.4942 \\
Misleading & 0/15 (0.00\%) & 13/15 (86.67\%) & 10/15 (66.67\%) & 10/15 (66.67\%) & 0.4519 \\
\hline
\end{tabular}
 \end{center}
\end{table}

\end{document}